\documentclass[twoside,leqno,twocolumn]{article}

\usepackage[letterpaper]{geometry}

\usepackage{ltexpprt}

\usepackage{bm}
\usepackage{amsfonts}
\usepackage{xcolor}
\usepackage{graphicx}
\usepackage{amsmath}
\usepackage[hidelinks]{hyperref}
\usepackage{algorithm}
\usepackage{algorithmic}
\usepackage[normalem]{ulem}

\newcommand{\circled}[1]{\raisebox{.5pt}{\textcircled{\raisebox{-.9pt} {#1}}}}

\begin{document}

\title{Verifying Tree Ensembles by Reasoning about Potential Instances}
\author{Laurens Devos\textsuperscript{*}
\and Wannes Meert\textsuperscript{*}
\and Jesse Davis\thanks{KU Leuven, Dept. of Computer Science; Leuven.AI, Belgium. \texttt{firstname.lastname@cs.kuleuven.be} }}

\date{}

\maketitle


\fancyfoot[R]{\scriptsize{Copyright \textcopyright\ 2021 by SIAM\\
Unauthorized reproduction of this article is prohibited}}





\begin{abstract} \small\baselineskip=9pt
Imagine being able to ask questions to a black box model such as “Which adversarial examples exist?”, “Does a specific attribute have a disproportionate effect on the model's prediction?” or “What kind of predictions could possibly be made for a partially described example?” This last question is particularly important if your partial description does not correspond to any observed example in your data, as it provides insight into how the model will extrapolate to unseen data. These capabilities would be extremely helpful as they would allow a user to better understand the model's behavior, particularly as it relates to issues such as robustness, fairness, and bias.  In this paper, we propose such an approach for an ensemble of trees.  Since, in general, this task is intractable we present a strategy that (1) can prune part of the input space given the question asked to simplify the problem; and (2) follows a divide and conquer approach that is incremental and can always return some answers and indicates which parts of the input domains are still uncertain. The usefulness of our approach is shown on a diverse set of use cases.
\end{abstract}

\section{Introduction}

Additive tree ensembles, which include random forests \cite{breiman01}
and gradient boosting trees (e.g.,~\cite{chen16,ke17,devos19}) are
powerful models that are widely used in practice. They have been
successfully applied to address challenging real-world tasks such as
providing rankings in web search~\cite{li2008mcrank}, valuing players' actions in
professional sports match to aid in player recruitment~\cite{decroos19}, modeling disease association~\cite{chen2018egbmmda}, and performing fault detection~\cite{zhang2018data}.
This wider adoption has come with an increased
awareness that it is crucial that we understand how the machine
learned model will behave, even in circumstances that we have not
encountered before.

One way to better understand a learned model's behavior, would be to
have the ability to ask questions such as:
\begin{itemize}
  \item[Q1] Given a correctly classified instance from the dataset, can
slightly perturbing it cause it to be misclassified?
\item[Q2] Given an instance, can we perturb its feature values such that the
model's predicted label corresponds to a specific class?
\item[Q3] Can changing the value of one particular attribute have a
disproportionate or unwanted effect on the model's prediction?
\item[Q4] Does there exist two instances such that the instances agree on
all attributes except for one where the model makes wildly different
predictions for their target value?
  \item[Q5] Given a partial description of an instance, can we find values
for the unknown attributes such that a certain label is predicted?
\end{itemize}
Clearly, knowing the answers to such questions is of great practical
importance: questions 1 and 2 are related to a model's \emph{robustness to
adversarial examples} whereas questions 3 and 4 are related to
\emph{fairness} and \emph{bias}. Finally, question 5 can give insight
into how the model will behave when forced to  \textit{extrapolate} into
areas of the instance-space that were not covered by the training data.

Given the obvious importance of these questions, there has been
significant interest in developing approaches that can answer
them~\cite{russell2019}. Unfortunately, existing work largely suffers from one
primary limitation: most approaches focus on solving one very specific
type of question. For example, there is a long line of work that
attempts to find adversarial examples using the  ${\infty}$-norm~\cite{einziger19,tornblom20,ranzato20,hchen19}.
While some of these systems are capable of dealing with any $p$-norm, they are evaluated using the $\infty$-norm.
One reason why this might be the case is that the $\infty$-norm is inherently easier than the others because different attributes do not interact; Wang et al. even showed that the complexity class of robustness checking can vary depending on the norm used for some models~\cite{wang20}.
However, it has been shown that in some cases, the $\infty$-norm is insufficient~\cite{schott2018,tramer2019,wang20}. Intuitively, the used norm should correspond to the human concept that is being modeled.
After all, an adversarial example is undesirable because, to a human, it is clear that the predicted class should remain unchanged with respect to the original example.
There are many adversarial examples that a human would consider close, but the $\infty$-norm does not.
Hence, verification tools should be flexible enough to model a wide range of concepts and should therefore not be limited to a single norm.

Additionally, there are targeted approaches for
evasion~\cite{kantchelian16}, providing explanations~\cite{ignatiev19},
or testing robustness and stability~\cite{ranzato20,tornblom20,hchen19}. 
Clearly, it is undesirable to have to design a separate approach to
answer to each question. Ideally, a verification tool should be flexible enough to solve a wide range of problems.

This paper attempts to fill this gap by developing a generic approach
that can provide answers to a broader class of questions about an
additive tree ensemble. Namely, we consider any question that can be
represented as a satisfiable modulo theory (SMT) formula. Because our
approach relies on theorem proving, which is intractable in general, naively 
feeding this to an SMT solver will not be an adequate solution. Therefore, we
develop two strategies for speeding up the process. Specifically, our
approach
(1) follows an incremental, divide-and-conquer strategy that can always
return some answers, and indicate which parts of the input domains are
still uncertain, and
(2)  prunes large parts of the model's input space given the question.
The divide-and-conquer strategy naturally decomposes the problem into
disjoint subproblems, and allows us to find multiple distinct satisfying
instances. Additionally, this information can be provided as feedback to the
user to potentially refine the question or update the background
knowledge to restrict the input space, making it an iterative procedure
with a human in the loop.
Empirically, we highlight the generality of our approach by providing a
diverse set of illustrative use cases.

\section{Preliminaries}

The framework described in this paper allow us to reason about \textbf{additive ensembles of binary trees}.\footnote{Note that all trees can be represented as binary trees.} A binary tree $T$ consists of nodes $n_i$ and has a special first node $n_0$ called the root node. There are two types of nodes. An \textit{internal node} $n$ stores a split condition defined on an input attribute and references to two child nodes $\mathrm{left}(n)$ and $\mathrm{right}(n)$. The split condition is either a less-than split $A_k < \tau$ defined on the real variable $A_k \in \mathcal{V}$ corresponding to a real attribute $A_k$ or a Boolean split defined on a variable corresponding to a Boolean attribute. A \textit{leaf node} is a node without children that stores an output value.

A tree is evaluated by recursively traversing it starting from the root node. For internal nodes, the node's split condition is tested; if the test succeeds, the procedure is recursively applied to the left child node, else, it is applied to the right child node. If a leaf node is encountered, the value stored in the leaf is returned and the procedure terminates.

An additive ensemble of trees is a sum of trees $\bm{T} = T_1 + \cdots + T_M$ and is evaluated by summing the evaluations of all trees. Specifically, the leaf values predicted by the individual trees are summed to obtain the final output of the ensemble. Examples of additive tree models are random forests with voting for classification or output averaging for regression, and gradient boosted trees. Both of these are powerful methods frequently used in practice.

We will be using \textbf{SMT} solvers (satisfiability modulo theories) to check logical theories.
SMT extends the Boolean satisfiability problem (often abbreviated SAT) with  a number of additional concepts. For this work, the most important capability is the addition of real variables and linear constraints between them. As SMT is more expressive than SAT, many SMT problems are of course intractable. However, powerful solvers like Z3~\cite{demoura08} are used to solve many real-world problems.

An SMT program is a logical theory defined on a set of \textbf{decision variables}. The theory defines constraints between the decision variables. A solver looks for an assignment to each of the decision variables such that the constraints defined in the theory hold. Either such assignment exists and the theory is \textbf{satisfiable}, or no such assignment exists and the model is unsatisfiable.
The assignments to the variables that satisfy the theory are called a \textbf{model} of the theory.
We will be using both real and Boolean decision variables.

Consider the following example SMT program defined on the real variable $R$ and Boolean variable $B$: $\neg B \lor (R > 5)$.
Examples of models of this theory are $\{B, R=9\}$ and $\{\neg B, R=3\}$.

\section{Problem Setting and Approach}

The problem setting we consider is the following:
\begin{description}
\item[Given:] An additive tree ensemble model, a question that can be represented as an SMT formula, and any available domain knowledge.
\item[Do:] Check if a (set of) instances exists that satisfies the requirements in the question.
\end{description}
Our approach is to translate the given information into a logical formula. An SMT solver can then be applied to check if a satisfying assignment to the formula exists. The SMT solver will either return \textit{\underline{yes}}, and provide a concrete instance that answers the question, or \textit{\underline{no}}, which  means that no instance exists that satisfies the provided information. Next we discuss encoding the ensemble, the question, and the optional background knowledge into a logical theorem as well as the computational complexity of this problem.

\subsection{Encoding the Ensemble}

Consider an additive tree ensemble trained on input data with $K$ attributes.
We first define the decision variables that correspond to the values used in the ensemble:
\begin{itemize}
    \item $K$ real or Boolean variables $\{A_k\}$ representing the $K$ input attributes,
    \item $M$ real tree output values $\{W_m\}$ corresponding to the outputs of the individual trees, and
    \item one real ensemble output variable $F$, which equals the sum the tree output values $W_1 + \cdots + W_M$.
\end{itemize}
The decision variables are collected in a set $\mathcal{V}$.

We then use the decision variables to translate a tree into a logical representation. A straightforward encoding is a disjunction of all root-to-leaf paths, where each root-to-leaf path is a conjunction of the conditions leading to that leaf. We use an equivalent but more compact encoding. Each tree $T_m$ is encoded by applying `$\mathrm{enc}$' to its root node and recursively encoding all its descendants.
An internal node $n$ is encoded as:
\begin{equation}
    \begin{array}{rl}
        \mathrm{enc}(n, \mathcal{V}) \rightarrow
        &(\ \  \mathrm{cond}(n) \land \mathrm{enc}(\mathrm{left}(n), \mathcal{V})\ ) \\
        \ \lor
        &( \neg\mathrm{cond}(n) \land \mathrm{enc}(\mathrm{right}(n), \mathcal{V})\ ),
    \end{array}
\end{equation}
where $\mathrm{left}(n)$ and $\mathrm{right}(n)$ are the left and right child nodes of the internal node, and $\mathrm{cond}(n)$ refers to the node's split condition.
Each internal node is a choice point: if the condition is true, you go left, else you go right. This is reflected in the encoding. Either the condition and the left branch's encoding are true, or the negated condition and the right branch's encoding are true.
A leaf node is encoded as:
\begin{equation}
    \mathrm{enc}(n, \mathcal{V}) \rightarrow (W_m = \mathrm{value}(n)),
\end{equation}
where $\mathrm{value}(n)$ is the output value stored in the leaf node.
This logical statement reflects that the output of the tree equals the value in the leaf, and this is only the case when all conditions leading to the leaf are true.

An additive ensemble $\bm{T} = \sum_m{T_m}$ is encoded as the conjunction of the encodings of the individual trees and an output constraint:
\begin{equation}
    \mathrm{enc}(\bm{T}, \mathcal{V}) \rightarrow 
    \left( \bigwedge_{m} \mathrm{enc}(T_m, \mathcal{V}) \right)
    \land 
    \left( F = \sum_{m} W_m \right),
\end{equation}
with $W_m$ the tree output variables in $\mathcal{V}$, $m = 1, \ldots, M$.

\begin{figure}
        \centering
        \def\svgwidth{5cm}
        \footnotesize 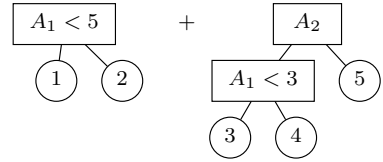
        \caption{An additive tree model}
        \label{fig:example1}
    \end{figure}

   The encoding of the ensemble in Figure~\ref{fig:example1} is:
    \begin{equation*}
        \begin{aligned}
            &[(A_1 < 5 \land W_1 = 1) \lor (A_1 \geq 5 \land W_1 = 2)] \\
            &\land \ [
                (A_2 \land [(A_1 < 3 \land W_2 = 3) \lor (A_1 \geq 3 \land W_2 = 4)]) \\
                &\lor \  (\neg A_2 \land W_2 = 5)] \\
            &\land \  F = W_1 + W_2.
        \end{aligned}
    \end{equation*}

    
\subsection{Encoding the Question}
The encoding $\mathrm{enc}(\bm{T}, \mathcal{V})$ of an additive tree ensemble $\bm{T}$ is only useful when combined with the encoding of a user-provided question.
Our approach works with any question that can be represented as a formula in the SMT-Lib~\cite{barrett16} language using (1) the variables in $\mathcal{V}$, and possibly (2) extra decision variables to formulate constraints.
We use $\mathrm{question}(\mathcal{V}, S)$ to denote the encoding of a question, with $S$ the set of additional decision variables.

A simple question about the ensemble in Figure~\ref{fig:example1} is
``Can an instance $x$ be constructed for which attribute 1 is less than 2 and the model's output is greater than 5?''
 This can be encoded as the following SMT formula:
        \begin{equation*}
            \mathrm{enc}(\bm{T}, \mathcal{V})
            \ \land \ (A_1 < 2) \ \land \ F > 5.
    \end{equation*}
It is also possible to add constraints between variables. For example, use $A_2 \Rightarrow (A_1 > 1)$ to add the constraint ``if the second attribute is positive, then the first attribute must be greater than 1.''

SMT also gives the flexilibility to ask more complicated questions, including ones that require reasoning about multiple instances at once. An example question reasoning about two instances of the ensemble in Figure~\ref{fig:example1} is:
``Find two instances $x$ and $x'$ that differ in exactly one attribute such that the model's prediction for $x$ is 2 units greater than its prediction for $x'$.''
Representing this question as an SMT formula requires introducing new Boolean decision variables $S_1$ and $S_2$, one for each attribute. We use primed variables to describe the second instance $x'$. The variables $F$ and $F'$ represent the output of the model for $x$ and $x'$ respectively. This question can be encoded as:

\begin{equation*}
        \begin{aligned}
            &\mathrm{enc}(\bm{T}, \mathcal{V}) \ \land \ \mathrm{enc}(\bm{T}, \mathcal{V}')\\
            &\quad\land [(\neg S_1 \land (A_1 = A_1')) \lor (S_1 \land (A_1 \neq A_1')] \\
            &\quad\land [(\neg S_2 \land (A_2 = A_2')) \lor (S_2 \land (A_2 \neq A_2')] \\
            &\quad\land [(S_1 \land \neg S_2) \lor (\neg S_1 \land S_2)] \\
            &\quad\land \ F = F' + 2.
        \end{aligned}
    \end{equation*}
A positive $S_1$ indicates that $x$ and $x'$ differ in the first attribute. A positive $S_2$ indicates $x$ and $x'$ differ in the second attribute. $S_1$ and $S_2$ must never be equal.

In general, when reasoning about multiple instances, we add an ensemble encoding using different variable sets $\mathcal{V}$ for each instance. Depending on the question, the variables in the different instances might be related. These relations must be encoded by constraints in $\mathrm{question}(\{\mathcal{V}\}, S)$, with $\{\mathcal{V}\}$ the set of all variable sets. There is no requirement for the ensemble to be the same for all instances. For example, it is possible to compare multiple 1-versus-all classifiers originating from a multi-class classifier.

\subsection{Background Knowledge}
The background knowledge $\mathit{BK}$ represents the implicit rules and constraints present in the dataset. Two sorts of background knowledge can be distinguished:
A first sort is knowledge about the problem domain. For example, in soccer, a cross must end in the opponent's penalty box. A second is feature engineering. For example, one-hot encoding usually requires an \textit{exactly-one} constraint, i.e., exactly one option is true, or when using 2-grams in a bag-of-words dataset, the truth of a 2-gram implies the truth of the individual words.
Generated instances that violate the background knowledge should be avoided, not only because these instances are not informative, but also because they might distort your results. E.g., exceedingly large outputs can be found by combining tree paths that are valid in a purely logical sense, but violate the background knowledge.

\subsection{Complexity Analysis}

Our setting means that answering a question about an ensemble boils down to checking if $\mathrm{enc}(\bm{T}, \mathcal{V}) \land \mathrm{question}(\mathcal{V}, S) \land \mathit{BK}$ is satisfiable.  Without considering the additional and arbitrarily complex constraints in $\mathrm{question}(\mathcal{V}, S)$, verifying whether $F > 0$ for a given additive ensemble is already NP-complete as it can be reduced from 3-SAT~\cite{kantchelian16}.

\section{Algorithm: Prune, Divide, and Conquer}

There are two main requirements:

\textit{Requirement 1}: The algorithm must be able to handle general questions. We cannot rely on question-specific optimizations used in prior work.
 
\textit{Requirement 2}: The algorithm must provide insights even if the question is intractable. The insights should inform the user about potential useful refinements to the question background knowledge.

Intuitively, the complexity of the problem can be seen as exponential in the number of leafs. We can understand this as follows: the ensemble output $F$ is defined as the sum of all tree outputs. The tree output is the value in the reached leaf node. The number of possible leaf-value combinations making up the sum $F$ is enormous. For example, for a small ensemble consisting of 10 trees each with 64 leafs would have $64^{10} \approx 10^{18}$ potential leaf-value combinations. This suggests we should somehow limit the number of leafs per tree.
Guided by this intuitive insight, we designed two strategies that reduce the number of leafs without changing the prediction made by the ensemble.

\subsection{Pruning Unreachable Branches}
The idea underlying this strategy is simple: the constraints in a question can make some branches in the trees inaccessible. For example, assume a question is about ``\textit{men aged 32 or older}''. 
If you reach a node splitting on $\mathrm{Age} < 23$, its left branch is never followed, because it contradicts the age condition in the question.

We can formulate this in more detail as follows. Consider the path from a node $n$ in a tree $T$ to the root node $n_0$. When moving from a child node to its parent node $p$, either a left or a right branch is followed. In the case of a left branch, the true condition $\mathrm{cond}(p)$ holds on the path from $n$ to $n_0$. If a right branch is taken, the negated condition $\neg \mathrm{cond}(p)$ holds.

Let $\mathrm{path}(n, T)$ be the conjunction of the (negated) conditions on the path from node $n$ to the root node of tree $T$. Given $\mathrm{question}(\mathcal{V}, S)$, if a tree branch rooted at $n$ is impossible given that question, i.e., $\mathrm{question}(\mathcal{V}, S) \land \mathrm{path}(n, T) \land \mathit{BK}$ does not have a solution, we can prune it from the model. Excluding the branch from the tree encoding $\mathrm{enc}(T, \mathcal{V})$ reduces the number of encoded leafs by the number of leafs below $n$.
As long as $\mathrm{question}(\mathcal{V}, S)$ is reasonably simple, this subproblem is much easier to solve than the full problem.

\subsection{Divide and Conquer: Split the Input Domain}
This strategy splits the input domain into two subdomains such that the number of still reachable leafs is maximally decreased. Consider Figure~\ref{fig:example1}. Initially, the domains of the input attributes $A_1$ and $A_2$ are $\mathbb{R}$ and $\{\mathit{True}, \mathit{False}\}$. If we split the domain of $A_1$ into $(-\infty, 5)$ and $[5, \infty)$, then node \circled{2} is unreachable in the first subdomain, and nodes \circled{1} and \circled{3} are unreachable in the second subdomain.

The procedure loops over all splits in the additive ensemble and counts the number of unreachable leafs in the first and second subdomains. It proceeds by splitting the input domain using the split with the highest unreachable leaf count. This produces two subproblems each of a reduced size. Each subproblem has an additional domain constraint, and the two domain constraints complement each other. Any satisfying solution of a subproblem is a satisfying solution of the original problem. If all subproblems are unsatisfiable, then the original problem is unsatisfiable.

\subsection{Combining Strategies: Prune, Divide and Conquer}

The two strategies above can be applied iteratively:
\begin{enumerate}
    \item\label{lbl:alg-prune} Prune the tree given the \textit{divide constraint} (initially unconstrained, i.e., \textit{True}) and the question.
    \item Apply the SMT solver on the pruned encoding of the trees and the question. If an answer is obtained within the timeout, report the answer; else stop the solver and continue to Step~\ref{lbl:alg-divide}.
    \item\label{lbl:alg-divide} Divide the input domain into two subdomains using the best split $C$ from the ensemble (e.g., $A_k < \tau$) and start two new problem instances at Step~\ref{lbl:alg-prune}. The \textit{divide constraint} is the current divide constraint appended by $C$ and $\neg C$ respectively.
\end{enumerate}

A major benefit of this approach is that the two subproblems produced by Step~\ref{lbl:alg-divide} can be solved independently in parallel, with only minor data moving and synchronization requirements. Our implementation can run the subproblems on a cluster of machines.

We can stop the algorithm at any point, even if we do not have an answer to our question. There are two ways in which we can interpret the partial results of the algorithm:
(1) The solver may have returned a \textit{yes} or \textit{no} answer to some subproblems. Depending the requirements, a single \textit{yes} answer with a generated instance might be enough. A \textit{no} indicates that no answer exists in a subdomain of the solution space. For example, when testing robustness, a \textit{no} indicates that robustness holds in the subdomain.
(2) The subproblems that have not been solved yet might indicate that finding a solution in this subspace is difficult. The more difficult the subspace, the more domain splits will have to be generated.
This may be informative to a domain expert: some subdomains may be uninteresting or (physically) impossible. Based on this information, the domain expert could reformulate the question or extend the background knowledge to prune some of these subdomains.
Pseudo-code is provided in Algorithm~\ref{alg:prune-and-split}.

\begin{algorithm}
\caption{The prune-divide-and-conquer algorithm}
\label{alg:prune-and-split}
\begin{algorithmic}[1] 
    \STATE \textit{stack} $\leftarrow$ $\{ \textit{unconstrained domain} \}$
    \WHILE{\textit{stack} not empty}
    \STATE \textit{domain} $\leftarrow$ pop from \textit{stack}
    \STATE \underline{prune} unreachable nodes in \textit{domain}
    \STATE \underline{solve} problem in \textit{domain}
    \IF {solution found}
    \STATE \textbf{report} solution
    \ELSIF {timeout}
    \STATE \textit{subdom1}, \textit{subdom2} $\leftarrow$ \underline{divide} \textit{domain}
    \STATE \textit{stack} $\leftarrow$ \textit{stack} $\cup$ $\{$ \textit{subdom1}, \textit{subdom2} $\}$
    \ENDIF
    \ENDWHILE
\end{algorithmic}
\end{algorithm}

\section{Use Cases}

All models are constructed by XGBoost~\cite{chen16}. We used XGBoost's early stopping functionality with a maximum number of trees of 50. The early stopping functionality stops the ensemble construction when no progress is made on the test set in the last 5 iterations. Our implementation is available from \href{https://github.com/laudv/treeck}{github.com/laudv/treeck}.

\subsection{Verifying Monotonicity}

A model is monotone in an attribute $A_{k^*}$ if an increase in the attribute's value results in an increase in the model's output. We can verify monotonicity as follows. Let $A_k$ and $A_k'$ be attribute variables and $F$ and $F'$ be ensemble outputs for a first and a second instance. Add the following constraints to the question encoding: $A_k = A_k'$ for all $k\neq k^*$, $A_{k^*}' < A_{k^*}'$, and $F > F'$. We can pass this question to our system.

If our system responds with a \textit{yes}, then we have two instances for which the monotonicity constraint is violated. If the system reports \textit{no}, then the monotonicity constraint holds.
We reproduced the simple synthetic dataset presented in the XGBoost documentation about the monotonicity feature.\footnote{\href{https://xgboost.readthedocs.io/en/latest/tutorials/monotonic.html}{xgboost.readthedocs.io/en/latest/tutorials/monotonic.html}}
We were able to verify that the models were indeed monotone.

\subsection{Generating Adversarial Examples}\label{sec:adv-ex}

This use case focuses on generating \textit{adversarial examples}, which corresponds to the following question:
\begin{quote}
    Given a correctly classified instance from the dataset, can a minor perturbation change the predicted output label to a desired value?
\end{quote}
This problem has been well studied for additive tree ensembles~\cite{kantchelian16,einziger19,hchen19} as well as for other algorithms (e.g. neural networks~\cite{szegedy13,goodfellow14,carlini17}, SVMs~\cite{biggio14}). However, existing approaches have formulated solution strategies specifically for this question. We illustrate that our generic approach can solve the same question.

We used an XGBoost model trained on the MNIST dataset that has 50 trees and a test set error rate of 2.66\%. The MNIST dataset consists of 70k grayscale handwritten digits of 28 by 28 pixels. The pixels are encoded as integer values between 0 and 255.

As most previous work, we will first use the $\infty$-norm to quantify the size of the perturbation and compare our method to the sound but incomplete multi-level approach by Chen et al.~\cite{hchen19}, the current state of the art. Their approach is incomplete because it sometimes outputs \textit{maybe SAT}, i.e., it cannot prove that no adversarial example exists, and it is unable to produce a concrete counter-example. When not stopped prematurely, our method is complete.

We randomly selected 250 digits from the dataset, 25 of each class. For a randomly selected digit $x$ with true label $y$, we pick a target label $y' \neq y$ and try to verify whether an $x'$ exists such that ${|| x - x' ||}_\infty < 10$ and $\bm{T}(x') = y'$. We repeat this for all target labels $y' \neq y$. This resulted in 2250 runs. In case no adversarial example $x'$ exists, our method is able to prove this for all cases, while the method of Chen et al. is not able to prove this in 5.7\% of the cases.

Wang et al. argue that the $\infty$-norm is not always sufficient: sometimes other distance functions are better suited to the application~\cite{wang20}. They show how Chen et al.'s approach can be extended to any $p$-norm. We now illustrate the flexibility of our method by going one step further and encoding a hybrid distance function:
For a given instance $x$, we look for an adversarial example $x'$ such that ${||x-x'||}_\infty < 75$ and ${||x-x'||}_1 < 3000$. That is, individual pixels can differ by at most 75 and there is a \textit{total budget} of 3000. Additionally, we constrain the model's confidence: it has to assign a probability of at most 1\% to the original label and at least 99\% to the adversarial label. We again select 250 digits resulting in 2250 runs, as before.

\begin{figure}
{
    \centering
    \def\svgwidth{\linewidth}
    \footnotesize 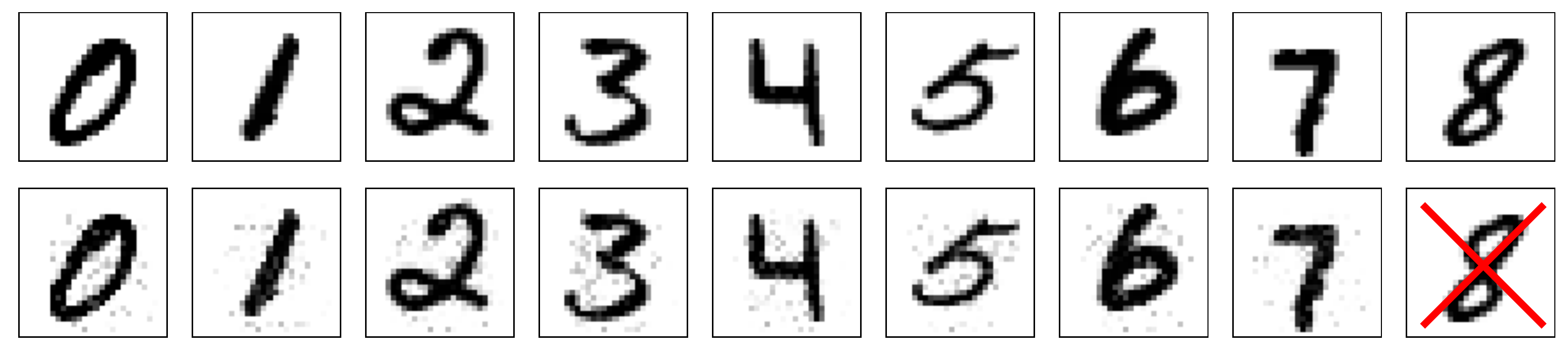

%
%
%
%
%
%
%

    \caption{Original and adversarial examples for MNIST generated by our approach. The top row shows the original instances which are correctly classified with high confidence. The bottom row shows the generated perturbed instances. For each of the original instances for zero through seven, it is possible to generate an adversarial example that is incorrectly classified as the number nine with high confidence. However, for the last digit, no adversarial exists, i.e., there is no perturbation that satisfies the given constraints such that the generated instance is classified as a nine.} 
\label{fig:mnist1}
}
\end{figure}

Figure~\ref{fig:mnist1} shows some examples. A few pixels can deviate by quite a lot (75), but the number of such pixels is limited by the total budget (3000). It is unlikely that a human would confuse any of these digits, certainly not with the same high confidence as the classifier.

Using this hybrid distance function, our method is more precise than Chen et al.'s method in 36.3\% of the cases. In these cases, our method is able to prove that no adversarial example exists while the competing method reports \textit{maybe SAT}.
Because Chen et al.'s method does not consider the additional constrains concerning total budget or the confidence of the model, it is not surprising that it is falsely optimistic about the existence of an adversarial example.
In the remaining 63.7\% of the cases, the methods report the same result, ignoring the fact that Chen et al.'s approach does not generate concrete counter-examples.

\begin{figure}
{
    \centering
    \def\svgwidth{\linewidth}
    \footnotesize 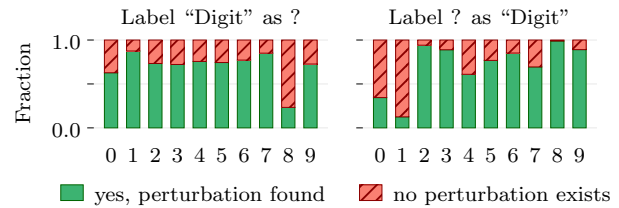
    \caption{The robustness as reported by our method of each MNIST digit averaged over 2250 targeted adversarial attacks. The left side shows how often the classifier could be \textit{fooled} into labeling the instance with label on the $x$-axis as another preselected digit. The right side shows the inverse: how frequently could the label of an instance be turned into the label on the $x$-axis.
    }
\label{fig:mnist2}
}
\end{figure}

We aggregate the results of our method in Figure~\ref{fig:mnist2}. The figure shows two quantities for each digit. The left side shows how difficult it is to change the label of each digit. The right side shows how difficult it is to perturb an instance such that the model predicts a specific target digit. We can see that it is hard to change an \textit{eight} into another digit, but it is easy to change any other digit into an \textit{eight}. The opposite is true for the \textit{one} digit, which is easy to change into any other digit, but it is hard to turn another digit into a \textit{one}.
The solver always terminates for this experiment and our approach takes between 10 and 45 seconds to answer the question. 

\subsection{Challenging Fairness}\label{sec:fairness}

\begin{figure}
	\begin{minipage}[t!]{0.4\linewidth}%
	\def\svgwidth{\linewidth}%
    \footnotesize 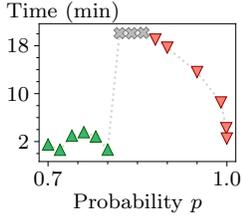%
    \end{minipage}%
    \begin{minipage}[t!]{0.55\linewidth}%
    \caption{A SAT phase transition for generating pairs of instances that differ only in a protected attribute, yet receive an opposite label. The confidence of the prediction $p$ is varied. The upwards and downwards triangles refer to \textit{yes} and \textit{no} responses. Crosses refer to timeouts.}%
    \label{fig:adult}%
   	\end{minipage}%
\end{figure}

Although our system cannot reason about \textit{individual fairness} as it was original defined~\cite{dwork12}, we can generate individuals that are treated unfairly. We use the Adult\footnote{\url{https://archive.ics.uci.edu/ml/datasets/adult}} dataset to illustrate this idea. This task is to predict whether an individual has a salary greater than 50k using information like age, education, race, sex, etc. We use an XGBoost model with 30 trees that achieves a test set  accuracy of 86.7\%. We ask the following question:
\begin{quote}
    Can we find two individuals A and B that only differ on one  protected attribute where the model makes a different prediction?
\end{quote}
We use sex as the protected attribute and search for pairs of examples where the model's predicted probability of earning less than 50k is $\geq p$ for individual A and $\leq 1-p$ for individual B. Figure~\ref{fig:adult} shows how varying $p$ affects the results. It is relatively easy to generate pairs instances that are treated unfairly when $p\leq 0.8.$ For $p> 0.85$, no such pairs exists. However, the solver does not terminate within 20 minutes when $p$ is between $0.8$ and $0.85$. This is referred to as the \textit{phase transition} of satisfiability problems~\cite{gent94}. 

\begin{figure}
	\begin{minipage}[t!]{0.4\linewidth}%
	\def\svgwidth{\linewidth}%
    \footnotesize 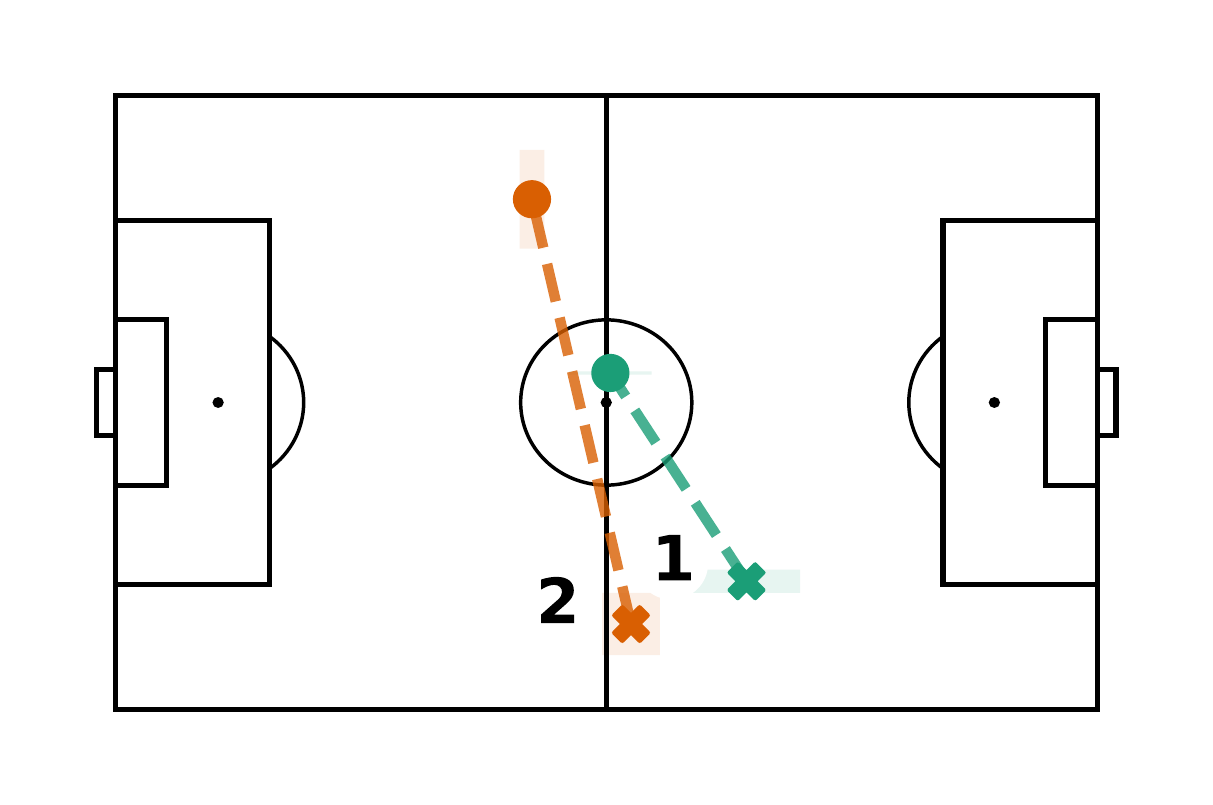%
    \end{minipage}%
    \begin{minipage}[t!]{0.55\linewidth}%
    \caption{Two midfield \textit{pass-dribble} sequences that have a probability greater than 10\%. The pass starts at the cross and ends at the circle. The dribble starts at the circle.}%
    \label{fig:soccer}
    \end{minipage}%
\end{figure}

\subsection{Detecting Dominant Attributes}\label{sec:dom-attr}

This use case attempts to predict the order of magnitude of YouTube\footnote{\url{https://www.kaggle.com/datasnaek/youtube-new}} video view counts (log 10) using a binary bag-of-words representation of the words appearing in the video's title and description. The XGBoost ensemble has 50 trees and achieves a test set  mean absolute error of 0.38. We address the following question:
\begin{quote}
  Is it possible to find a set of words and a single additional word such that the predicted view count for the set with and without the additional word differs by two order of magnitude?
\end{quote}
This questions contrasts with the previous two use cases as we neither \textit{investigate a particular instance (Section~\ref{sec:adv-ex})} nor \textit{focus on one specific protected attribute (Section~\ref{sec:fairness})}. The task is to find \emph{any} attribute that, when flipped, causes a significant change in the predicted value.

This produces some interesting results. We constrained the result to have less than 12 words.  A bag-of-words representation does not enforce a word order, so we re-ordered the words to make a sentence. We also added filler words in non-bold. These words do not appear in the bag-of-words representation and therefore have no effect on the predicted view count.

The first example is constrained to contain the words \textit{night}, \textit{talk}, and \textit{show}. This is one of the results:
\begin{quote}
    \textbf{Night talk show video}: \textbf{pop drama} about the \textbf{latest hot Christmas house album} (\textbf{\underline{remix}}).
\end{quote}
Without the word \textit{remix}, this title is predicted to receive 200,000 views. However, if \textit{remix} is included, the predicted view count rises to 30 million.
A second example of a video title containing the words: \textit{news}, \textit{breaking}, and \textit{channel}:
\begin{quote}
    \textbf{Breaking news channel}: \textbf{\underline{no}} \textbf{weird money vlogs today challenge} (\textbf{full movie}).
\end{quote}
Without the word \textit{no}, this video is predicted to receive 100,000 views. With \textit{no}, it is predicted to receive 100 million views.
A final example is:
\begin{quote}
    The \textbf{12 avengers challenge} \textbf{Paul}, the \textbf{\underline{Christmas} pop fashion king} in \textbf{DE}.
\end{quote}
When the word \textit{Christmas} is omitted, the video is predicted to get 1 million views. However, if \textit{Christmas} is included, the prediction drops to  only 1000 views.

While the stakes are not high for view count prediction, similar situations can arise in prediction tasks of insurance companies, law enforcement, the health care sector, etc. The robustness of the above boosted tree ensemble is clearly inadequate for such much more sensitive applications.

\subsection{Querying the Model}

Our final use case involves analyzing real-world event stream from professional soccer matches. The machine learning task is to estimate the probability of scoring a goal in the near future (e.g., within the next ten actions) from a particular game state.\footnote{\url{https://github.com/ML-KULeuven/socceraction}} This enables valuing on-the-ball actions, which is a crucial task for soccer analytics~\cite{decroos19}. We trained an XGBoost model with 50 trees using 1.1 million actions over multiple games. The attributes are: action type (e.g. pass, shot, throw in, penalty, etc.), $x$ and $y$ coordinate of action, body part used (foot, head, other), current goal scores, and time remaining.
There are many domain constraints (e.g., a penalty cannot be executed using your hands), and it is difficult to list them all. In this use case, the (partial) results from the system can be used to refine the background knowledge.
We consider the following question:
\begin{quote}
    Can we find a two-action sequence involving a backward pass in the midfield that results in a game state with a probability greater than 10\% of scoring in the near future?\footnote{Goals are exceedingly rare in soccer, and very few game states would have such a high probability.}
\end{quote}
This is a relevant question as the soccer analytics community is interested in understanding the usefulness of backwards passes far away from the goal. Our method proved that \textit{pass-pass} sequences in the midfield cannot have a probability greater than 10\%. When also allowing dribbles, the system generated several \textit{pass-dribble} sequences, two of which are shown in Figure~\ref{fig:soccer}. Intuitively, these sequences could represent valuable situations because the backward pass could simultaneously switch the direction of play and get the ball to a player who has space where he can advance the ball.  

\section{Algorithm Analysis}

\begin{figure}
    \centering
    \def\svgwidth{\linewidth}
    \footnotesize 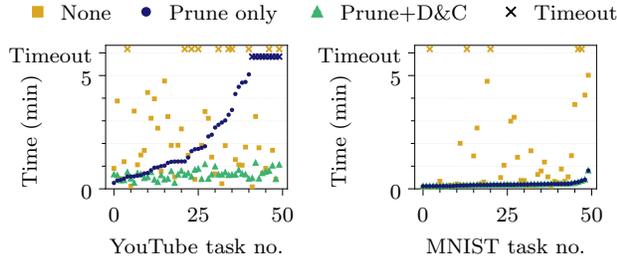
    \caption{Timings of 50 YouTube tasks on the left, and 50 MNIST tasks on the right. Timings are shown for the following three cases: passing the full encoding directly to the SMT solver (``None''), using only the pruning step (``Prune only''), and using the pruning and divide-and-conquer algorithm (``Prune+D\&C''). }
    \label{fig:youtube_time}
\end{figure}

To evaluate effectiveness of our prune plus divide and conquer algorithm, we compare three variants: (1) the full encoding is passed directly to the SMT solver, (2) applying the pruning step before passing the encoding to the SMT solver, and (3) applying both the pruning and divide-and-conquer steps. We consider two question types. On YouTube, we randomly picked eight fixed words and asked the same question as in the use case. On MNIST, we randomly picked an instance and an incorrect adversarial target label.

Figure~\ref{fig:youtube_time} shows timings on 50 random task for each data set. For the YouTube tasks on the left, all timeouts are eliminated when the full algorithm is active. For the MNIST tasks on the right, the pruning step is the most effective. This makes sense: the adversarial example question is heavily constrained; each pixel can only deviate by $\delta$, which makes many branches in the trees unreachable. The number of leafs before pruning is about 5000 and the pruning eliminates 80\% of them on average. The SMT solver can solve this reduced problem easily, so no divide-and-conquer is necessary.

\section{Related Work}

For tree ensembles, most related work has focused on finding~\cite{einziger19} and evading~\cite{kantchelian16} adversarial examples. The former also uses an SMT solver, while the latter uses mixed-integer linear programming. In contrast to the work in these papers, we have presented a method to perform targeted attacks: we do not just change the label, we change it to a specific value. More generally, we can handle a wider range of questions that do not necessarily reason about one specific instance from the dataset, a setting inherent to the \textit{adversarial question}.

Other work has moved beyond individual adversarial examples and proposed methods to prove \textit{stability} and \textit{robustness} of additive tree ensembles. Ranzato and Zanella \cite{ranzato20} propose a method that uses a similar prune and divide-and-conquer approach, but they do not use an SMT solver but an approach specifically tuned for the \textit{stability} problem. T\"ornblom and Nadjm-Tehrani \cite{tornblom20} use a technique they call \textit{equivalence class partitioning} that enumerates all possible outputs of the model. This approach does not scale, however, for problems with more attributes like MNIST. The current state of the art in robustness checking is by Chen et al.~\cite{hchen19}, work that has been expanded upon by \cite{wang20}. They use a graph representation and a merging strategy to compute an upper bound on the ensembles output.

Ignatiev et al.\@ \cite{ignatiev19} have used an SMT encoding of boosted trees in a method to provide global explanations and validate heuristic explanations. Their focus is, therefore, also limited to a single question. 

\section{Conclusion}

We presented an approach that answers general questions about additive tree ensembles. Our approach applies theorem proving using an SMT solver, which is intractable in general. We propose a prune, divide and conquer algorithm that (1) speeds up the computation, and (2) provides partial results when the full computation takes too long. To illustrate the abilities of our approach, we provide a diverse set of use cases and experiments.
We show that we can solve 5.7\% more problems than the current state of the art on the task the competitor was designed for. Additionally, we can solve a wide range of questions whereas existing systems focus on a single question.

\section*{Acknowledgements}

LD is supported by KU Leuven Research Fund (C14/17/070) and Research Foundation-Flanders (FWO). This research received funding from Research Foundation-Flanders under EOS No. 3099257.

\bibliographystyle{siam}
\bibliography{references}

\end{document}